\documentclass{article}

\usepackage{microtype}
\usepackage{graphicx}
\usepackage{subfigure}
\usepackage{booktabs} % for professional tables
\usepackage{tikz}
\usetikzlibrary{arrows.meta, positioning, calc, shapes.geometric}
\usepackage[preprint]{icml2026}
\usepackage{amsmath}
\usepackage{amssymb}
\usepackage{mathtools}
\usepackage{amsthm}

\usepackage[capitalize,noabbrev]{cleveref}

\theoremstyle{plain}

\theoremstyle{definition}

\theoremstyle{remark}

\usepackage[textsize=tiny]{todonotes}

\icmltitlerunning{title}

\begin{document}

\twocolumn[
% \icmltitle{TorRicc: Global--Local Geometry Proxies for Diagnosing OOD Robustness}
\icmltitle{Representation Geometry as a Diagnostic for Out-of-Distribution Robustness}



\icmlsetsymbol{equal}{*}

\begin{icmlauthorlist}
\icmlauthor{Farid Hazratian}{equal,sch}
\icmlauthor{Dr. Ali Zia}{equal,yyy}
\end{icmlauthorlist}

\icmlaffiliation{sch}{School of Mathematics, Computer Science and Statistics, University of Tehran, Tehran, Iran}
\icmlaffiliation{yyy}{School of Computing, Engineering \& Mathematical Sciences, La Trobe University , Melbourne, Australia }

\icmlcorrespondingauthor{Farid Hazratian}{faridish96@gmail.com}
\icmlcorrespondingauthor{Ali Zia}{A.Zia@latrobe.edu.au}

\icmlkeywords{Machine Learning, ICML}

\vskip 0.3in
]



\printAffiliationsAndNotice{\icmlEqualContribution} % otherwise use the standard text.

\begin{abstract}

Robust generalization under distribution shift remains difficult to monitor and optimize in the absence of target-domain labels, as models with similar in-distribution accuracy can exhibit markedly different out-of-distribution (OOD) performance. While prior work has focused on training-time regularization and low-order representation statistics, little is known about whether the geometric structure of learned embeddings provides reliable post-hoc signals of robustness. We propose a geometry-based diagnostic framework that constructs class-conditional mutual $k$-nearest-neighbor graphs from in-distribution embeddings and extracts two complementary invariants: a global spectral complexity proxy based on the reduced log-determinant of the normalized Laplacian, and a local smoothness measure based on Ollivier--Ricci curvature. Across multiple architectures, training regimes, and corruption benchmarks, we find that lower spectral complexity and higher mean curvature consistently predict stronger OOD accuracy across checkpoints. Controlled perturbations and topological analyses further show that these signals reflect meaningful representation structure rather than superficial embedding statistics. Our results demonstrate that representation geometry enables interpretable, label-free robustness diagnosis and supports reliable unsupervised checkpoint selection under distribution shift.

\end{abstract}

\section{Introduction}
\label{sec:intro}

Modern deep networks routinely achieve near-perfect accuracy on their training and validation distributions, yet their behaviour under distribution shift remains highly unstable. Models with comparable in-distribution performance may exhibit substantially different out-of-distribution (OOD) accuracy, and these differences often vary across architectures, optimization regimes, and even across training checkpoints \cite{arjovsky2019irm,rame2022fishr,cha2021swad}. In practical settings, where target-domain labels are unavailable, this creates a fundamental challenge: robustness is difficult to monitor, compare, and select for, rendering early stopping, checkpoint selection, and model deployment under shift largely heuristic.

Recent empirical evidence suggests that robustness differences are reflected in the geometric and topological structure of learned representations \cite{rieck2018neural,guss2018characterizing,wang2020understanding}. However, most training objectives and post-hoc diagnostics rely on low-order similarity measures, such as feature norms, covariance spectra, or representation similarity scores (e.g., CKA) \cite{kornblith2019similarity,jiang2019fantastic}. While computationally convenient, these quantities collapse high-dimensional geometry into coarse global statistics and are largely insensitive to local regularity, higher-order connectivity, and class-conditional structure. Consequently, we currently lack principled and reproducible tools for diagnosing robustness directly from trained representations.

Graph-based spectral analysis and topological data analysis have long been used to probe data geometry and learning dynamics, including Laplacian eigenvalues, heat kernels, entropy and log-determinant complexity measures \cite{chung1997spectral,vonluxburg2007tutorial,belkin2003laplacian,smola2003kernels}, as well as persistent homology summaries \cite{rieck2018neural,guss2018characterizing,barannikov2022representation}. While these tools provide valuable descriptive insights into connectivity and multiscale structure, they are rarely evaluated as quantitative predictors of robustness or generalization, and their practical utility for model selection remains unclear.

In this work, we do not propose a new training objective, regularizer, or domain generalization algorithm. Instead, we study the problem of \emph{checkpoint-level, label-free, post-hoc robustness diagnosis and selection}: can geometric properties computed from in-distribution (ID) embeddings alone serve as reliable indicators of OOD performance? Our focus is on extracting structural signals from representations that are predictive of robustness without access to target-domain data or labels.

To this end, we propose \textsc{TorRicc}, a geometry-based diagnostic pipeline that lifts learned embeddings to class-conditional mutual $k$-nearest-neighbor graphs and extracts two complementary invariants. First, we introduce a global spectral complexity proxy inspired by analytic torsion, computed as the reduced log-determinant of the normalized Laplacian. Unlike single-eigenvalue or entropy-based descriptors, this quantity aggregates the full spectrum and admits a combinatorial interpretation through the Matrix--Tree theorem \cite{raysinger1971torsion,lyons2005determinantal}. Second, we measure local smoothness using Ollivier--Ricci curvature, which quantifies neighborhood contraction under optimal transport \cite{ollivier2007ricci}. Together, these signals characterize global redundancy and local stability of class manifolds in representation space. We additionally report compact persistent homology summaries as secondary baselines \cite{rieck2018neural,barannikov2022representation}.

Empirically, we evaluate \textsc{TorRicc} on checkpoints of ResNet-18 and ViT-S/16 models trained on CIFAR-10 under ERM and contrastive objectives, and tested on CIFAR-10.1, CIFAR-10.2, CIFAR-10-C, and Tiny-ImageNet-C. We observe consistent trends: lower spectral complexity and higher mean curvature are strongly associated with improved OOD accuracy across corruption types and severities. These correlations deteriorate under structure-destroying controls, such as label or feature shuffling and degree-preserving rewiring, and remain stable under isometry-preserving transformations, such as random projections. This behavior indicates that the signals capture task-aligned geometric organization rather than numerical artifacts.

Our results suggest that representation geometry provides actionable information for robustness monitoring and model selection. In particular, geometric invariants derived from ID embeddings enable reliable unsupervised ranking of checkpoints, approaching oracle selection without access to OOD labels. This supports practical deployment scenarios in which robustness must be assessed under limited supervision. Our contributions are as follows:
% \paragraph{Contributions.}
\begin{itemize}
    \item \textbf{Geometry-based post-hoc robustness diagnostics:}
    Addressing the lack of principled label-free tools for robustness monitoring, we introduce \textsc{TorRicc}, a checkpoint-level framework that constructs class-conditional mutual $k$-NN graphs from ID embeddings and extracts global (torsion-inspired spectral complexity) and local (Ollivier--Ricci curvature) geometric invariants.

    \item \textbf{Structural indicators of OOD robustness:}
    Filling the gap between descriptive representation analysis and predictive robustness assessment, we show that reduced spectral complexity and increased local curvature are consistent, interpretable predictors of OOD performance. Controlled perturbations and ablations demonstrate that these signals arise from nontrivial representation geometry rather than superficial statistics.

    \item \textbf{Unsupervised checkpoint selection under shift:}
    Overcoming the reliance on target-domain labels for model selection, we demonstrate that the proposed geometric signals enable reliable, architecture-agnostic ranking and selection of robust checkpoints, achieving near-oracle performance and supporting practical early stopping and model comparison.
\end{itemize}

\begin{figure*}[t]
\centering
\includegraphics[width=0.96\linewidth]{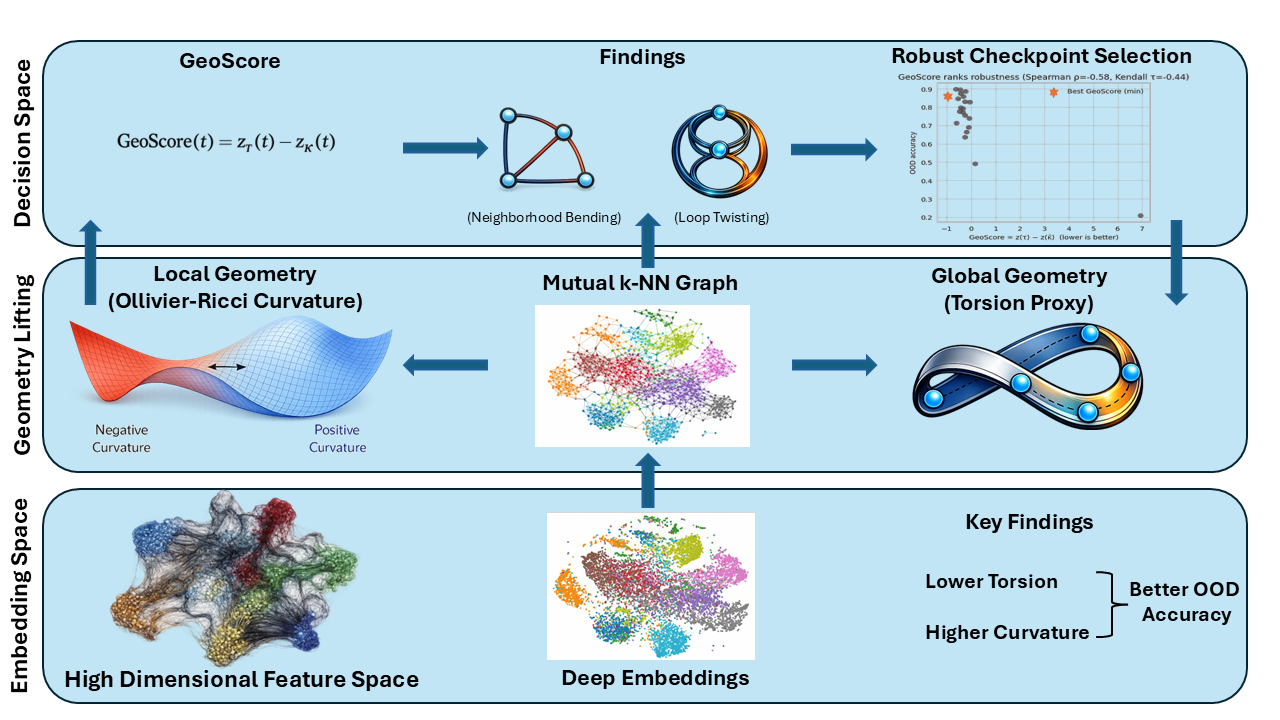}
\caption{
Overview of the proposed \emph{TorRicc} pipeline for post-hoc robustness diagnosis.
Given in-distribution embeddings from a trained checkpoint, we construct class-conditional mutual $k$-NN graphs and extract complementary global (torsion-inspired spectral complexity) and local (Ollivier--Ricci curvature) invariants.
These geometry-based signals are fused into a lightweight score that enables unsupervised checkpoint ranking and robustness monitoring under distribution shift.
}

\label{fig:pipeline}
\end{figure*}

%%%%%%%%%%%%%%%%%%%%%%%%%%%%%%%%%%%%%%%%%%%%%%%%%%%%%%%%%%%%%%%%%%%%%%%%%%%%%%%%
%%%%%%%%%%%%%%%%%%%%%%%%%%%%%%%%%%%%%%%%%%%%%%%%%%%%%%%%%%%%%%%%%%%%%%%%%%%%%%%%

\section{Related Work}
\label{sec:related}

% \paragraph{Domain generalization and OOD robustness.}
Robust generalization under distribution shift has been widely studied through domain generalization and invariant representation learning \cite{arjovsky2019irm,ahuja2020invariant,krueger2021out}. Other approaches emphasize optimization and regularization strategies that promote flat or stable solutions, including Fishr \cite{rame2022fishr}, SWAD \cite{cha2021swad}, and stochastic weight averaging \cite{izmailov2018averaging}. These methods primarily operate at training time and require access to multiple environments or carefully designed objectives. In contrast, our work focuses on post-hoc, label-free robustness diagnosis from learned representations and enables checkpoint selection without modifying training procedures.

% \paragraph{Representation similarity and standard diagnostics.}
A large literature studies representation comparison and characterization using similarity measures and global statistics, including CKA and related kernel-based metrics \cite{kornblith2019similarity,nguyen2021do}, as well as covariance spectra, feature norms, and gradient-based measures \cite{jiang2019fantastic,foret2020sharpness}. Recent work has further examined representation collapse and anisotropy in self-supervised learning \cite{wang2020understanding}. While computationally efficient, these diagnostics primarily capture low-order or global properties and are often insensitive to local neighborhood structure and class-conditional geometry. Our approach instead targets higher-order connectivity and local regularity in representation space.

% \paragraph{Spectral and topological analysis of representations.}
Spectral methods and topological data analysis have long been used to probe data geometry and learning dynamics, including Laplacian-based descriptors, heat kernels, and persistent homology \cite{belkin2003laplacian,smola2003kernels,rieck2018neural,guss2018characterizing,barannikov2022representation}. These tools provide valuable descriptive insights into connectivity, cycles, and multiscale structure. However, prior work has largely treated such quantities as explanatory or exploratory, rather than as quantitative predictors of robustness or generalization. We systematically evaluate spectral and topological invariants as predictive signals for OOD performance.

% \paragraph{Discrete curvature in learning systems.}
Discrete notions of Ricci curvature, particularly Ollivier--Ricci curvature, translate geometric contraction and expansion to graph domains \cite{ollivier2007ricci}. Recent studies have applied curvature to analyze information bottlenecks and over-squashing in graph neural networks \cite{topping2022oversquashing}. These works focus on message passing on explicit graph-structured data. In contrast, we study curvature on graphs induced by latent representations of vision models and relate it to robustness under distribution shift.

% \paragraph{Summary.}
To our knowledge, no prior work has combined analytic torsion--inspired spectral complexity and discrete curvature to diagnose robustness in deep representation spaces. By evaluating these geometric invariants alongside standard diagnostics and demonstrating their utility for unsupervised checkpoint selection, our work establishes representation geometry as a practical and previously unexplored tool for robustness monitoring.

\section{TorRicc: Global--Local Geometry Proxies}
\label{sec:torricc}
We formulate checkpoint-level robustness diagnosis as a mapping from in-distribution embeddings to geometric summaries that capture both global connectivity and local stability of class manifolds. Given a trained model checkpoint, \textsc{TorRicc} extracts embeddings, constructs class-conditional mutual $k$-NN graphs, and computes complementary global and local invariants without access to target-domain labels. These quantities are subsequently fused into a lightweight geometry-based score for within-run checkpoint ranking. Figure~\ref{fig:pipeline} illustrates the overall TorRicc pipeline.

\subsection{Embeddings and graph construction}
Given a checkpoint $\theta$ and dataset $\mathcal{D}=\{(x_i,y_i)\}$, we extract embeddings $z_i=f_\theta(x_i)\in\mathbb{R}^d$ from a fixed intermediate layer. All embeddings are $\ell_2$-normalized to remove scale effects, and class-balanced subsampling is used to control graph size and prevent class imbalance from dominating spectral statistics. Additional preprocessing details are provided in Appendix~\ref{app:preprocess}.

To isolate intra-class geometric structure, we construct separate graphs for each class. For each class $c$, we build a mutual $k$-NN graph $G_c=(V_c,E_c,W_c)$, where an undirected edge connects $i,j\in V_c$ if each lies among the $k$ nearest neighbors of the other. The mutual criterion suppresses asymmetric and noisy neighbor relations, yielding more stable local connectivity.

Edge weights are assigned using a self-tuning Gaussian kernel
\[
w_{ij}=\exp\!\Big(-\frac{\|z_i-z_j\|^2}{\sigma_i\sigma_j}\Big),
\]
where $\sigma_i$ denotes the distance to the $k$-th neighbor of $i$. This adaptive weighting reduces sensitivity to local density variations and produces sparse, scale-invariant affinity graphs that faithfully capture class-conditional manifold geometry.

\subsection{Global spectral complexity: Torsion proxy}
Let $W$ denote the weighted adjacency matrix of $G_c$ and $L_{\mathrm{sym}}=I-D^{-1/2}WD^{-1/2}$ its normalized Laplacian. We define the torsion proxy as the reduced log-determinant
\[
\tau(G_c)=\log\det\nolimits^*(L_{\mathrm{sym}}),
\]
where $\det^*$ excludes zero eigenvalues.

Unlike single-eigenvalue descriptors (e.g., $\lambda_2$) or entropy-based measures, $\tau(G_c)$ aggregates the full nonzero Laplacian spectrum and admits a principled combinatorial interpretation. By Kirchhoff’s Matrix--Tree theorem, $\exp(\tau(G_c))$ is proportional to the number of spanning trees in $G_c$, reflecting global connectivity redundancy and cyclic complexity.

We average $\tau(G_c)$ across classes to obtain $\mathcal{T}(\theta)$. Lower values indicate simpler, more coherent class manifolds with fewer unstable pathways, while higher values correspond to highly entangled, cycle-rich structures. Empirically, we find that reduced spectral complexity is consistently associated with improved OOD robustness.

\subsection{Local regularity: Ollivier--Ricci curvature}
To quantify local smoothness of representation manifolds, we compute Ollivier--Ricci (OR) curvature on each class-conditional graph $G_c$. For each node $u$, we define a neighbor distribution
\[
\mu_u(x)=\frac{w_{ux}}{\sum_{z\in\mathcal{N}(u)}w_{uz}},
\]
which assigns probability mass to adjacent nodes proportional to edge weights.

For an edge $(u,v)$, OR curvature is defined as
\[
\kappa(u,v)=1-\frac{W_1(\mu_u,\mu_v)}{d(u,v)},
\]
where $W_1$ denotes the Wasserstein-1 distance and $d(u,v)$ is the graph distance. We approximate $W_1$ using entropic regularization for computational efficiency.

We summarize each class by its mean edge curvature
\[
\bar{\kappa}(G_c)=\frac{1}{|E_c|}\sum_{(u,v)\in E_c}\kappa(u,v),
\]
and average across classes to obtain $\bar{\kappa}(\theta)$. Higher curvature reflects stronger neighborhood overlap and local contraction, indicating smoother and more stable representation geometry. Empirically, we observe that increased mean curvature correlates with improved OOD robustness.

\subsection{Unified checkpoint ranking: GeoScore}
Let $\tau(\theta)$ and $\bar{\kappa}(\theta)$ denote the class-averaged torsion and curvature for checkpoint $\theta$. Since these quantities may differ in scale across runs and architectures, we $z$-normalize each metric across checkpoints within the same training run. We then define
\[
\mathrm{GeoScore}(\theta)=z_\tau(\theta)-z_\kappa(\theta).
\]

This linear combination reflects the complementary roles of global spectral regularity and local smoothness: lower torsion and higher curvature are both associated with improved robustness. The sign convention ensures that lower GeoScore values correspond to more favorable geometric structure.

GeoScore is intended as a lightweight, unsupervised ranking criterion for comparing checkpoints within a run, rather than as an absolute predictor of OOD accuracy across unrelated models.

\subsection{Theoretical perspective}
Our diagnostics admit an interpretable connection to stability, margin-based generalization, and representation smoothness. Robust classification under distribution shift requires that small input perturbations induce limited displacement in representation space relative to class boundaries, which is closely related to Lipschitz continuity and effective margin size.

Positive Ollivier--Ricci curvature reflects local contraction of neighborhood distributions under optimal transport, promoting stability of nearest-neighbor relations. Conversely, low torsion indicates reduced global redundancy and fewer unstable pathways in class manifolds. Together, these properties imply smoother and more globally regularized representations.

This perspective is intended as theoretical intuition rather than a formal guarantee, and helps explain the empirical effectiveness of the proposed diagnostics. Extended discussion is provided in Appendix~\ref{app:theory}.

%%%%%%%%%%%%%%%%%%%%%%%%%%%%%%%%%%%%%%%%%%%%%%%%%%%%%%%%%%%%%%%%%%%%%%%%%%%%%%%%%%%%%%%%%%%%%%%%%%%%%%%%%%%%%%%%%%%%%%%%%%%%%%%%%%%%%%%%%%%%%%%%%%%%%%%%%%%%%%%%%%%%%%%%%%%%%%%%
\section{Experiments}
\label{sec:experiments}

\subsection{Goals and evaluation questions}
\label{sec:exp_goals}
We study whether \emph{representation geometry computed from ID embeddings only} can diagnose robustness under distribution shift at the \emph{checkpoint level}. Our evaluation focuses on
(i) \textbf{predictivity:} do torsion and curvature track OOD accuracy across checkpoints?;
(ii) \textbf{stress testing:} do trends strengthen with shift severity and disappear under structure-breaking controls?;
(iii) \textbf{practicality:} can we select robust checkpoints without any OOD labels?;
and (iv) \textbf{generality:} do findings transfer across architectures and datasets.

Together, these questions frame our study as an empirical investigation into geometry as a predictive signal for robustness, rather than a post-hoc descriptive tool.

\subsection{Setup}
\label{sec:exp_setup}

\paragraph{Datasets and shifts:}
We train on CIFAR-10 and evaluate on CIFAR-10.1, CIFAR-10.2, and CIFAR-10-C (severity 1--5). To test cross-dataset generality, we additionally evaluate on Tiny-ImageNet-C (\S\ref{sec:tinyc}).

\paragraph{Models and checkpoints:}
Unless otherwise stated, we use ResNet-18 trained with ERM under standard augmentation and regularization, saving checkpoints regularly across training. Key conclusions are replicated on ViT-S/16 (Appendix).

\paragraph{Source-only diagnostic protocol:}
For each checkpoint $\theta_t$, we extract ID embeddings and compute all TorRicc metrics using the class-conditional mutual $k$-NN graphs described in \S\ref{sec:torricc}. We then evaluate true OOD accuracy on the shifted benchmarks and quantify association across checkpoints using Spearman rank correlation $\rho$ (Kendall $\tau$ in Appendix). This is strictly \emph{label-free and source-only}: no target labels are used for diagnostics or selection.

\subsection{Baseline robustness under shift}
\label{sec:ood_baselines}

We first verify that the evaluation benchmarks induce non-trivial distribution shift and that robustness is not inherited by a frozen representation.

\begin{table}[t]
\centering
\small
\begin{tabular}{lccc}
\toprule
Dataset & Acc & CI$_{\text{lo}}$ & CI$_{\text{hi}}$ \\
\midrule
\multicolumn{4}{l}{\textbf{ERM}} \\
CIFAR-10 (test)            & 0.9533 & 0.9400 & 0.9633 \\
CIFAR-10.1 v6              & 0.8853 & 0.8680 & 0.9040 \\
CIFAR-10.2                 & 0.8220 & 0.8060 & 0.8380 \\
CIFAR-10-C (sev=5)         & 0.7643 & 0.7557 & 0.7749 \\
\midrule
\multicolumn{4}{l}{\textbf{Probe}} \\
CIFAR-10 (test)            & 0.7953 & 0.7767 & 0.8147 \\
CIFAR-10.1 v6              & 0.6387 & 0.6107 & 0.6640 \\
CIFAR-10.2                 & 0.6387 & 0.6140 & 0.6620 \\
CIFAR-10-C (sev=5)         & 0.4178 & 0.4068 & 0.4282 \\
\bottomrule
\end{tabular}
\caption{OOD accuracies with confidence intervals.}
\label{tab:erm_probe_ood}
\end{table}

Table~\ref{tab:erm_probe_ood} shows that ERM degrades gracefully under shift, whereas frozen features collapse. This confirms that robust geometry must be learned end-to-end, motivating checkpoint-level analysis.

\subsection{Main result: geometry predicts OOD accuracy across checkpoints}
\label{sec:main_corr}

We next test whether global spectral complexity (torsion proxy) and local regularity (mean Ollivier--Ricci curvature) track OOD accuracy across checkpoints.

\begin{figure}[t]
\centering
\includegraphics[width=0.9\linewidth]{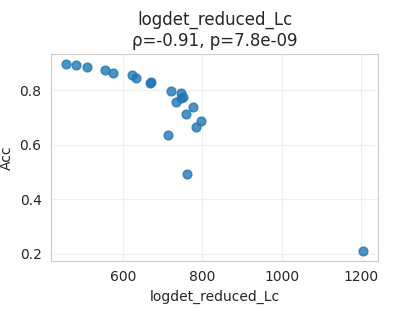}
\caption{Torsion proxy vs.\ OOD accuracy on CIFAR-10.1. Lower torsion correlates with higher robustness.}
\label{fig:torsion_ood}
\end{figure}

\begin{figure}[t]
\centering
\includegraphics[width=0.9\linewidth]{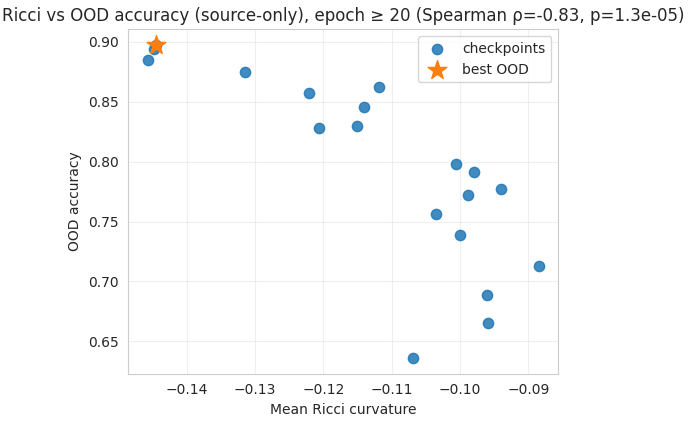}
\caption{Mean Ollivier--Ricci curvature vs.\ OOD accuracy. Higher curvature correlates with better robustness.}
\label{fig:ricci_ood}
\end{figure}

On CIFAR-family shifts, we observe consistent monotonic trends: robust checkpoints exhibit \emph{lower} torsion and \emph{higher} mean curvature (Fig.~\ref{fig:torsion_ood}, Fig.~\ref{fig:ricci_ood}). These relationships persist throughout training rather than emerging only at convergence.

\begin{table}[t]
\centering
% \resizebox{\columnwidth}{!}{%
\begin{tabular}{lcc}
\toprule
Metric & Spearman $\rho$ & $p$-value \\
\midrule
Torsion (logdet)        & $-0.88$ & $1.4\!\times\!10^{-15}$ \\
Ricci curvature        & $+0.68$ & $8.6\!\times\!10^{-4}$ \\
Heat trace             & $+0.84$ & $1.9\!\times\!10^{-6}$ \\
PH $H_0$ lifetime      & $-0.70$ & $3.2\!\times\!10^{-4}$ \\
\midrule
Anisotropy             & $+0.93$ & $3.6\!\times\!10^{-15}$ \\
CKA                    & $+0.11$ & $2.1\!\times\!10^{-1}$ \\
Feature norm           & $-0.91$ & $7.8\!\times\!10^{-16}$ \\
\bottomrule
\end{tabular}
% }
\caption{Correlation comparison with standard diagnostics.}
\label{tab:correlation_comparison}
\end{table}

Table~\ref{tab:correlation_comparison} shows that TorRicc metrics are among the strongest predictors, while also remaining interpretable in terms of graph geometry.

\subsection{Harder shifts amplify the signal: CIFAR-10-C severity sweep}
\label{sec:severity}

\begin{figure}[t]
\centering
\includegraphics[width=0.9\linewidth]{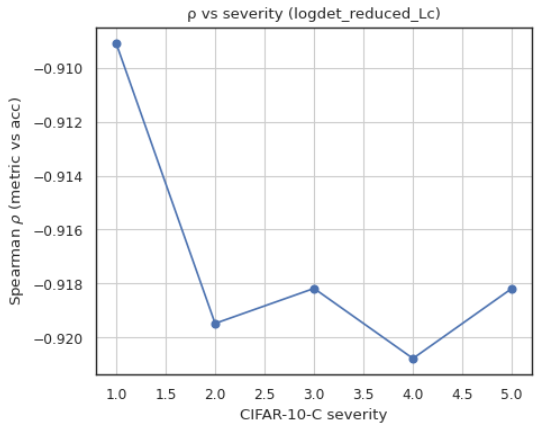}
\caption{Correlation vs.\ CIFAR-10-C severity.}
\label{fig:severity_sweep}
\end{figure}

Robustness diagnostics should become increasingly informative as distribution shifts intensify. We test this hypothesis by evaluating checkpoint-level correlations between geometric metrics and OOD accuracy across CIFAR-10-C corruption severities 1–5.

Figure~\ref{fig:severity_sweep} reports the Spearman correlation between each diagnostic and accuracy as a function of severity. As severity increases, the magnitude of correlation for both torsion and mean curvature grows monotonically. In particular, torsion exhibits progressively stronger negative correlation, while curvature shows increasingly strong positive correlation.
This trend indicates that geometric organization becomes increasingly predictive as distribution shifts intensify. Under mild perturbations, most checkpoints retain similar class structure, limiting discriminability, whereas under severe shifts unstable or over-entangled manifolds deteriorate more rapidly, amplifying the contrast captured by torsion and curvature. This supports our hypothesis that representation geometry primarily reflects robustness under challenging shifts rather than in-distribution performance.

\subsection{Sanity checks: structure-breaking controls}
\label{sec:controls_main}

\begin{figure*}[t]
\centering
\includegraphics[width=\linewidth]{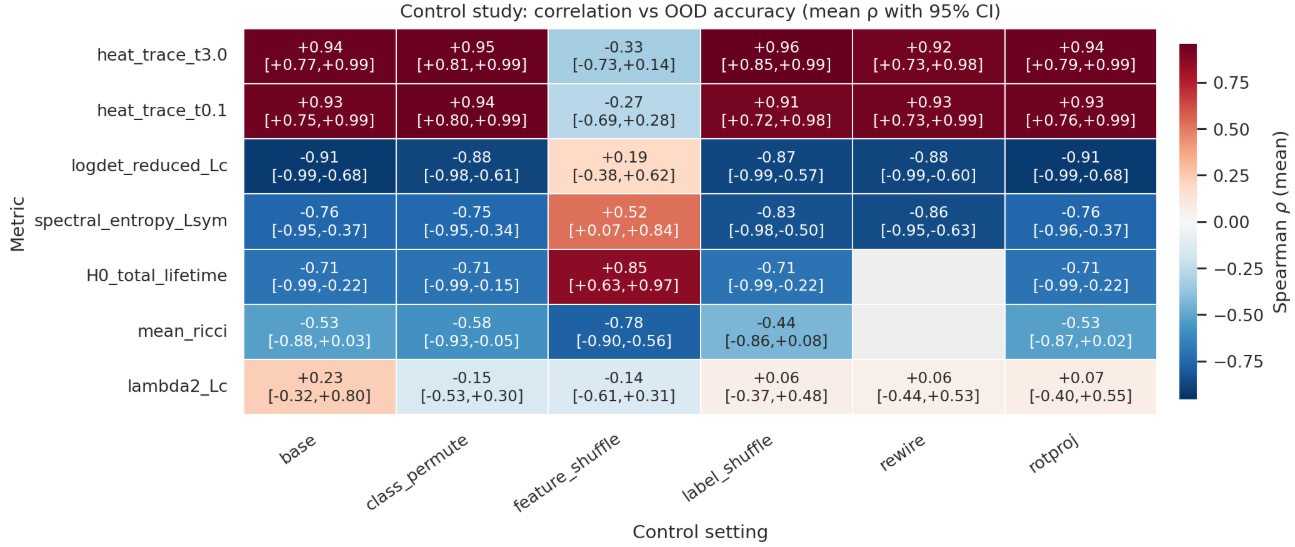}
\caption{Control study under structure-breaking interventions.}
\label{fig:controls_topology}
\end{figure*}

Correlations alone do not guarantee that diagnostics capture meaningful structure. They may arise from confounding factors such as scale, degree distribution, or feature variance. We therefore perform controlled interventions that disrupt semantic and geometric organization prior to graph construction.

Specifically, we consider label shuffling, feature shuffling, and degree-preserving rewiring. These operations destroy class-aligned neighborhood structure while approximately preserving low-level statistics. Figure~\ref{fig:controls_topology} summarizes the resulting correlations across metrics.

Under all structure-breaking controls, the correlations of torsion and curvature with OOD accuracy collapse or invert. In contrast, isometry-preserving transformations (orthogonal rotations and projections) largely preserve the signal, as shown in the Appendix heatmap. These findings indicate that TorRicc does not rely on trivial graph properties, but depends critically on task-aligned neighborhood geometry. The collapse under rewiring further rules out explanations based solely on degree or density effects.

\subsection{Practical utility: unsupervised checkpoint selection}
\label{sec:ckpt_selection_main}
Beyond correlation analysis, we evaluate whether geometric diagnostics can support practical model selection without access to target labels. Given a sequence of checkpoints from a training run, we select models using torsion, curvature, or GeoScore alone.

Table~\ref{tab:checkpoint_selection} reports the resulting OOD accuracy compared with oracle selection based on true target performance.
Both torsion- and GeoScore-based selection recover near-oracle checkpoints, substantially outperforming random or early-epoch selection.
Notably, curvature-only selection performs less consistently in isolation, reflecting its sensitivity to local fluctuations. GeoScore mitigates this by integrating global and local information, yielding more stable rankings. These results demonstrate that representation geometry can be operationalized for early stopping and model comparison in settings where validation on target domains is infeasible.

\begin{table}[ht]
\centering
\small
\begin{tabular}{lccc}
\toprule
Selector & Epoch & Ckpt & OOD Acc \\
\midrule
Oracle          & 199 & ep199 & 0.8975 \\
Torsion-only    & 199 & ep199 & 0.8975 \\
Curvature-only  &  60 & ep060 & 0.7125 \\
GeoScore        & 160 & ep160 & 0.8625 \\
\bottomrule
\end{tabular}
\caption{Unsupervised checkpoint selection (CIFAR).}
\label{tab:checkpoint_selection}
\end{table}
% Table~\ref{tab:checkpoint_selection} demonstrates that geometry-driven selection recovers near-oracle models without access to OOD labels.

\subsection{Joint geometry view across checkpoints}
\label{sec:joint_geometry_main}

\begin{figure}[t]
\centering
\includegraphics[width=0.9\linewidth]{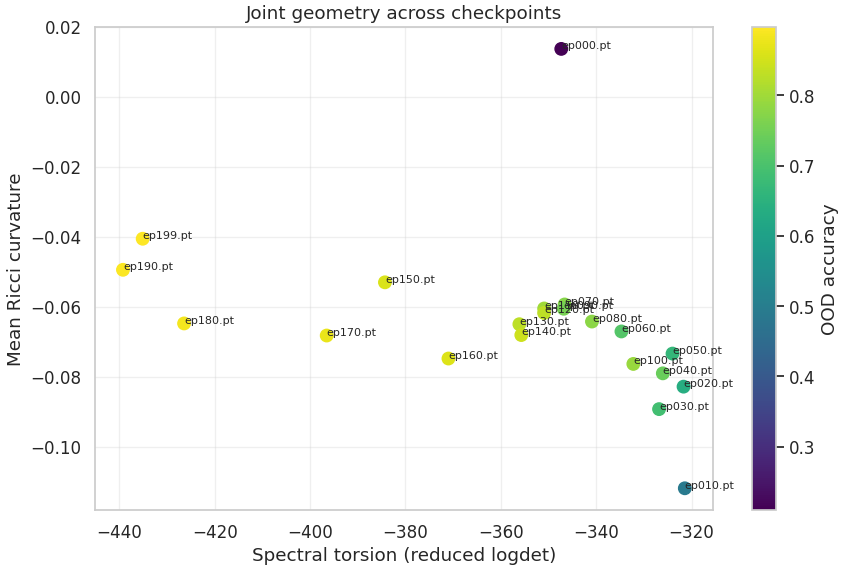}
\caption{Joint torsion--curvature plane across checkpoints.}
\label{fig:joint_geometry}
\end{figure}

Single-number diagnostics can hide failure modes, because robustness can be limited either
by global manifold ``tangling'' (captured by torsion) or by local instability of neighborhoods
(captured by curvature). We therefore visualize checkpoints in the two-dimensional
$(\tau,\bar{\kappa})$ plane, coloring each point by OOD accuracy.

Figure~\ref{fig:joint_geometry} reveals a coherent robustness regime: high-accuracy
checkpoints concentrate in the low-torsion / high-curvature region, and training trajectories
tend to move toward this region as robustness improves. Importantly, the joint plot clarifies
why single-metric selection can fail: some checkpoints achieve relatively low torsion but still
exhibit strongly negative curvature (poor local contraction), while others have improved
curvature but remain globally complex. These mixed cases motivate GeoScore as a simple
global--local fusion for within-run ranking, while preserving interpretability through the
two-axis decomposition.

\subsection{Cross-dataset validation: Tiny-ImageNet-C}
\label{sec:tinyc}

\begin{table}[t]
\centering
\small
\begin{tabular}{lcc}
\toprule
Metric & Spearman $\rho$ & $p$-value \\
\midrule
Torsion proxy    & $-0.774$ & $6.1\!\times\!10^{-5}$ \\
Mean curvature   & $+0.847$ & $2.5\!\times\!10^{-6}$ \\
GeoScore         & $-0.832$ & $5.5\!\times\!10^{-6}$ \\
Heat trace       & $+0.714$ & $4.0\!\times\!10^{-4}$ \\
PH $H_0$         & $-0.729$ & $2.6\!\times\!10^{-4}$ \\
\bottomrule
\end{tabular}
\caption{Tiny-ImageNet-C correlations (source-only).}
\label{tab:tiny_corr}
\end{table}

\begin{figure}[t]
\centering
\includegraphics[width=0.9\linewidth]{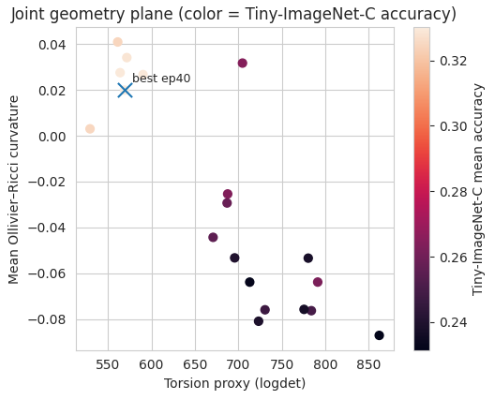}
\caption{Curvature vs.\ Tiny-ImageNet-C accuracy.}
\label{fig:tiny_ricci}
\end{figure}

\begin{figure}[t]
\centering
\includegraphics[width=0.9\linewidth]{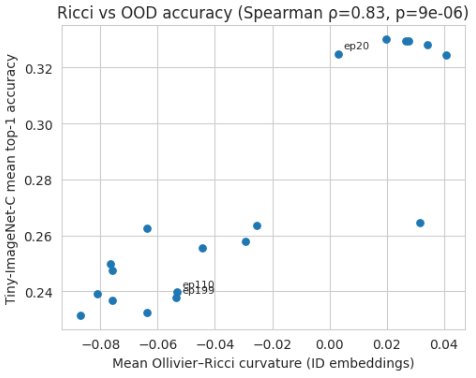}
\caption{Joint geometry on Tiny-ImageNet-C.}
\label{fig:tiny_joint}
\end{figure}

\begin{table}[t]
\centering
\small
\begin{tabular}{lccc}
\toprule
Selector & Epoch & Ckpt & Acc \\
\midrule
Oracle        & 40 & ep040 & 0.3301 \\
Torsion-only  & 20 & ep020 & 0.3249 \\
Curvature     & 60 & ep060 & 0.3246 \\
GeoScore      & 60 & ep060 & 0.3246 \\
\bottomrule
\end{tabular}
\caption{Tiny-ImageNet-C checkpoint selection.}
\label{tab:tiny_ckpt_selection}
\end{table}

To test whether TorRicc transfers beyond CIFAR-family shifts, we evaluate the \emph{same}
checkpoints under a cross-dataset shift using Tiny-ImageNet-C. This setting is harsher than
CIFAR-10-C in two ways: it introduces (i) distributional shift via corruptions and (ii)
semantic mismatch relative to the CIFAR-10 training distribution. We keep the diagnostic
strictly source-only: all metrics are computed from CIFAR-10 ID embeddings, and Tiny-ImageNet-C
labels are used only for reporting accuracy.

Table~\ref{tab:tiny_corr} reports strong checkpoint-level associations under this shift:
mean curvature correlates positively with Tiny-ImageNet-C accuracy, while torsion and GeoScore
correlate negatively, with statistically significant $p$-values. Table~\ref{tab:tiny_ckpt_selection}
shows that geometry-based selection remains near-oracle torsion- and GeoScore-based choices select checkpoints whose Tiny-ImageNet-C accuracy is close to the best observed checkpoint.

% Figures~\ref{fig:tiny_ricci} and \ref{fig:tiny_joint} provide complementary intuition.
Figure~\ref{fig:tiny_ricci} shows that curvature computed on ID embeddings tracks cross-dataset
robustness monotonically. Figure~\ref{fig:tiny_joint} recovers the same geometric regime seen
on CIFAR: robust checkpoints occupy the low-torsion/high-curvature area. Taken together, these
results suggest TorRicc is not overfitting to a particular corruption benchmark; instead it
captures representation organization that remains predictive under a qualitatively different
evaluation distribution.

We next assess the sensitivity of our conclusions to design choices in graph construction and embedding preprocessing. We vary the neighborhood size $k$, feature extraction layer (avgpool vs.\ pre-classifier), and preprocessing strategy (with and without PCA whitening).

Table~\ref{tab:ablations} reports a representative sweep. Across these settings, the qualitative relationship between geometry and robustness remains stable: lower torsion and higher curvature are consistently associated with improved OOD performance. Although absolute metric values vary with $k$ and layer depth, the relative trends and checkpoint ordering are preserved. Detailed per-class and correlation analyses are provided in Appendix~\ref{app:ablations}. This indicates that TorRicc reflects robust structural properties under reasonable design variations.

\begin{table}[ht]
\centering
\caption{Layer and $k$ sweep.}
\label{tab:ablations}
\resizebox{\columnwidth}{!}{%
\begin{tabular}{llrrrrrr}
\toprule
Layer & $k$ &
$\log\det^* L$ & Entropy & $\lambda_2$ & $h(0.1)$ & $H_0$ & $\bar{\kappa}$ \\
\midrule
avgpool & 5  & -235.4 & 4.98 & $3.4\!\times\!10^{-16}$ & 251.9 & 468.2 &  0.042 \\
avgpool & 10 & 498.7  & 5.30 & $1.3\!\times\!10^{-16}$ & 247.6 & 468.2 & -0.111 \\
avgpool & 15 & 1002.2 & 5.37 & $3.7\!\times\!10^{-16}$ & 245.0 & 468.2 & -0.133 \\
\bottomrule
\end{tabular}%
}
\end{table}

% Results are stable across neighborhood size, layer choice, and preprocessing.

\subsection{Additional geometry baselines}
\label{sec:extra_baselines_main}

We compare TorRicc with alternative geometry-aware descriptors, including multiscale heat traces and persistent homology (PH) summaries. As reported in Appendices~\ref{app:heat} and~\ref{app:ph} results , heat traces provide moderately strong correlations across diffusion scales, reflecting global connectivity patterns. However, they require careful scale tuning and lack the combinatorial interpretability of torsion. PH-based summaries capture coarse topological features but exhibit weaker and less stable predictive power. In particular, $H_0$ lifetimes correlate inconsistently under severe shifts, and $H_1$ features are sensitive to sampling noise. 
These results show that weighted spectral and curvature-based metrics capture robustness-relevant structure more reliably than topology-only or single-eigenvalue descriptors, combining interpretability with stable predictive performance.

%%%%%%%%%%%%%%%%%%%%%%%%%%%%%%%%%%%%%%%%%%%%%%%%%%%%%%%%%%%%%%%%%%%%%%%%%%%%%%%%%%%%%%%%%%%%%%%%%%%%%%%%%%%%%%%%%%%%%%%%%%%%%%%%%%%%%%%%%%%%%%%%%%%%%%%%%%%%%%%%%%%%%%%%%%%%%%%%

\section{Conclusion}
\label{sec:con}
We showed that global spectral complexity and local curvature of class-conditional representation graphs provide reliable, label-free indicators of robustness under distribution shift. Across architectures, checkpoints, and shift types, lower torsion and higher Ollivier--Ricci curvature consistently predict stronger OOD performance. These geometry-based diagnostics enable near-oracle checkpoint selection using in-distribution data only, and capture robustness-relevant structure that is not reflected by standard similarity or norm-based measures.
Our analysis is diagnostic rather than causal and is currently evaluated on vision benchmarks, leaving extensions to other modalities and settings as an important direction for future work. Our results nevertheless suggest that representation geometry offers a principled foundation for monitoring and improving robustness in modern deep learning systems.

\bibliography{references}

@article{arjovsky2019irm,
  title={Invariant risk minimization},
  author={Arjovsky, Martin and Bottou, L{\'e}on and Gulrajani, Ishaan and Lopez-Paz, David},
  journal={arXiv preprint arXiv:1907.02893},
  year={2019}
}

@inproceedings{rame2022fishr,
  title={Fishr: Invariant gradient variances for out-of-distribution generalization},
  author={Rame, Alexandre and Dancette, Corentin and Cord, Matthieu},
  booktitle={International Conference on Machine Learning},
  pages={18347--18377},
  year={2022},
  organization={PMLR}
}

@article{cha2021swad,
  title={Swad: Domain generalization by seeking flat minima},
  author={Cha, Junbum and Chun, Sanghyuk and Lee, Kyungjae and Cho, Han-Cheol and Park, Seunghyun and Lee, Yunsung and Park, Sungrae},
  journal={Advances in Neural Information Processing Systems},
  volume={34},
  pages={22405--22418},
  year={2021}
}

@inproceedings{kornblith2019similarity,
  title={Similarity of neural network representations revisited},
  author={Kornblith, Simon and Norouzi, Mohammad and Lee, Honglak and Hinton, Geoffrey},
  booktitle={International conference on machine learning},
  pages={3519--3529},
  year={2019},
  organization={PMlR}
}

@inproceedings{ahuja2020invariant,
  title={Invariant risk minimization games},
  author={Ahuja, Kartik and Shanmugam, Karthikeyan and Varshney, Kush and Dhurandhar, Amit},
  booktitle={International Conference on Machine Learning},
  pages={145--155},
  year={2020},
  organization={PMLR}
}

@InProceedings{krueger2021out,
  title = 	 {Out-of-Distribution Generalization via Risk Extrapolation (REx)},
  author =       {Krueger, David and Caballero, Ethan and Jacobsen, Joern-Henrik and Zhang, Amy and Binas, Jonathan and Zhang, Dinghuai and Priol, Remi Le and Courville, Aaron},
  booktitle = 	 {Proceedings of the 38th International Conference on Machine Learning},
  pages = 	 {5815--5826},
  year = 	 {2021},
  editor = 	 {Meila, Marina and Zhang, Tong},
  volume = 	 {139},
  series = 	 {Proceedings of Machine Learning Research},
  month = 	 {18--24 Jul},
  publisher =    {PMLR}
}

@article{izmailov2018averaging,
  title={Averaging weights leads to wider optima and better generalization},
  author={Izmailov, Pavel and Podoprikhin, Dmitrii and Garipov, Timur and Vetrov, Dmitry and Wilson, Andrew Gordon},
  journal={arXiv preprint arXiv:1803.05407},
  year={2018}
}

@article{nguyen2021do,
  title={Do wide and deep networks learn the same things? uncovering how neural network representations vary with width and depth},
  author={Nguyen, Thao and Raghu, Maithra and Kornblith, Simon},
  journal={arXiv preprint arXiv:2010.15327},
  year={2020}
}

@article{jiang2019fantastic,
  title={Fantastic generalization measures and where to find them},
  author={Jiang, Yiding and Neyshabur, Behnam and Mobahi, Hossein and Krishnan, Dilip and Bengio, Samy},
  journal={arXiv preprint arXiv:1912.02178},
  year={2019}
}

@article{foret2020sharpness,
  title={Sharpness-aware minimization for efficiently improving generalization},
  author={Foret, Pierre and Kleiner, Ariel and Mobahi, Hossein and Neyshabur, Behnam},
  journal={arXiv preprint arXiv:2010.01412},
  year={2020}
}

@inproceedings{wang2020understanding,
  title={Understanding contrastive representation learning through alignment and uniformity on the hypersphere},
  author={Wang, Tongzhou and Isola, Phillip},
  booktitle={International conference on machine learning},
  pages={9929--9939},
  year={2020},
  organization={PMLR}
}

@article{belkin2003laplacian,
  title={Laplacian eigenmaps for dimensionality reduction and data representation},
  author={Belkin, Mikhail and Niyogi, Partha},
  journal={Neural computation},
  volume={15},
  number={6},
  pages={1373--1396},
  year={2003},
  publisher={MIT Press}
}

@article{rieck2018neural,
  title={Neural persistence: A complexity measure for deep neural networks using algebraic topology},
  author={Rieck, Bastian and Togninalli, Matteo and Bock, Christian and Moor, Michael and Horn, Max and Gumbsch, Thomas and Borgwardt, Karsten},
  journal={arXiv preprint arXiv:1812.09764},
  year={2018}
}

@article{ollivier2007ricci,
  title={Ricci curvature of metric spaces},
  author={Ollivier, Yann},
  journal={Comptes Rendus Mathematique},
  volume={345},
  number={11},
  pages={643--646},
  year={2007},
  publisher={Elsevier}
}

@article{guss2018characterizing,
  title={On characterizing the capacity of neural networks using algebraic topology},
  author={Guss, William H and Salakhutdinov, Ruslan},
  journal={arXiv preprint arXiv:1802.04443},
  year={2018}
}

@inproceedings{smola2003kernels,
  title={Kernels and Regularization on Graphs},
  author={Smola, Alexander J. and Kondor, Risi},
  booktitle={Computational Learning Theory and Kernel Machines (COLT/Kernel 2003)},
  series={Lecture Notes in Computer Science},
  volume={2777},
  pages={144--158},
  year={2003},
  publisher={Springer}
}

@inproceedings{topping2022oversquashing,
  title={Understanding over-squashing and bottlenecks on graphs via curvature},
  author={Topping, Jake and Di Giovanni, Francesco and Chamberlain, Benjamin P. and Dong, Xiaowen and Bronstein, Michael M.},
  booktitle={International Conference on Learning Representations (ICLR)},
  year={2022}
}

@inproceedings{barannikov2022representation,
  title={Representation Topology Divergence: A Method for Comparing Neural Network Representations},
  author={Barannikov, Serguei and Trofimov, Ilya and Balabin, Nikita and Burnaev, Evgeny},
  booktitle={Proceedings of the 39th International Conference on Machine Learning (ICML)},
  volume={162},
  pages={1607--1626},
  year={2022},
  publisher={PMLR}
}

@book{chung1997spectral,
  title={Spectral Graph Theory},
  author={Chung, Fan},
  year={1997},
  publisher={AMS}
}

@article{vonluxburg2007tutorial,
  title={A tutorial on spectral clustering},
  author={von Luxburg, Ulrike},
  journal={Statistics and Computing},
  year={2007}
}

@article{raysinger1971torsion,
  title={R-torsion and the Laplacian},
  author={Ray, Daniel and Singer, Isadore},
  journal={Advances in Mathematics},
  year={1971}
}

@article{lyons2005determinantal,
  title={Determinantal probability measures},
  author={Lyons, Russell},
  journal={Publ. Math. IHES},
  year={2005}
}
\bibliographystyle{icml2025}

%%%%%%%%%%%%%%%%%%%%%%%%%%%%%%%%%%%%%%%%%%%%%%%%%%%%%%%%%%%%%%%%%%%%%%%%%%%%%%%
%%%%%%%%%%%%%%%%%%%%%%%%%%%%%%%%%%%%%%%%%%%%%%%%%%%%%%%%%%%%%%%%%%%%%%%%%%%%%%%
% APPENDIX
%%%%%%%%%%%%%%%%%%%%%%%%%%%%%%%%%%%%%%%%%%%%%%%%%%%%%%%%%%%%%%%%%%%%%%%%%%%%%%%
%%%%%%%%%%%%%%%%%%%%%%%%%%%%%%%%%%%%%%%%%%%%%%%%%%%%%%%%%%%%%%%%%%%%%%%%%%%%%%%
\newpage
\appendix
\onecolumn
% \section{Appendix.}

% \appendix
\section{ Implementation Details and Embedding Preprocessing}
\label{app:preprocess}

All experiments are implemented in PyTorch. Embeddings are extracted using forward hooks at fixed network layers. Nearest-neighbor search is performed using FAISS. Wasserstein distances for curvature computation are estimated using entropic regularization.

Experiments are conducted on NVIDIA A100 GPUs. Typical graph construction and metric computation require 5--15 minutes per checkpoint depending on dataset size. Hyperparameters and software versions are provided in the supplementary material.

All embedding vectors are $\ell_2$-normalized prior to graph construction. In some analyses, we additionally apply PCA whitening to reduce global anisotropy and prevent dominant variance directions from distorting distance and spectral statistics. Specifically, embeddings are projected onto their principal components and rescaled to unit variance.

We observe that the main qualitative trends reported in the paper are robust to this preprocessing choice. Results with and without PCA whitening are reported in Appendix~\ref{app:additional}.

\section{Persistent Homology Analysis}
\label{app:ph}
Graph-based spectral methods and topological data analysis have long been used to probe the structure of data manifolds and learning dynamics. For example, Laplacian eigenvalues, heat kernels, and log-determinant complexity measures have been used to characterize data geometry \cite{belkin2003laplacian,smola2003kernels}. In parallel, topological data analysis (TDA), particularly persistent homology, has been applied to probe neural network training and representation evolution \cite{rieck2018neural,guss2018characterizing,barannikov2022representation}. These approaches provide valuable descriptive insights into connectivity, cycles, and multiscale structure of learned manifolds. However, prior work typically treats these spectral or topological quantities as descriptive tools rather than \emph{quantitative predictors} of generalization. Our contribution is to systematically evaluate such descriptors by correlating them with OOD performance and by demonstrating their practical utility (e.g. for early stopping). This goes beyond existing studies by treating geometry-derived metrics as \emph{predictive signals} rather than post-hoc explanations.

\begin{figure*}[t]
\centering
\includegraphics[width=0.9\linewidth]{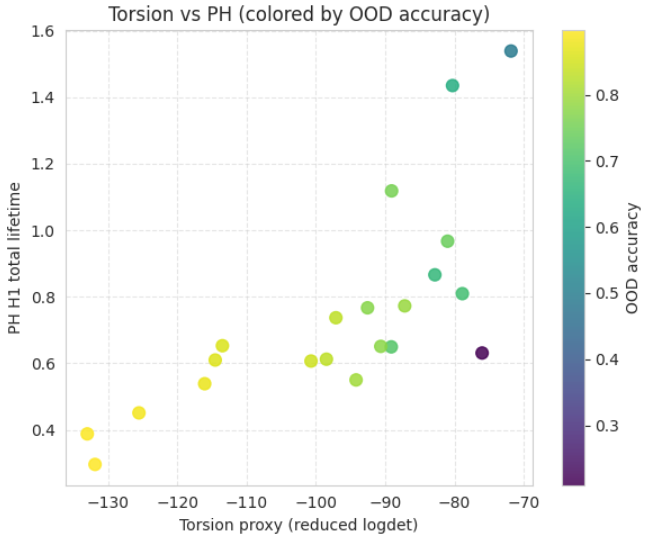}
\caption{Torsion proxy vs OOD accuracy.}
\label{fig:ph_ood}
\end{figure*}

As a purely topological measure (ignoring edge weights and exact distances), we compute persistent homology (PH) on the point cloud of embeddings as a baseline descriptor. We focus on 0-dimensional and 1-dimensional homological features to capture the merging of clusters and the presence of loops, respectively. For each class, we consider its set of normalized embeddings (optionally projected to a lower-dimensional subspace for efficiency) and construct a Vietoris–Rips filtration: we start with each point as an isolated component at radius $r=0$, then gradually increase $r$, adding an edge between any two points when the distance $\|z_i - z_j\|$ falls below $r$. As $r$ grows, connected components merge (this is recorded as death of a $H_0$ feature) and loops are formed when a cycle of edges appears (birth of an $H_1$ feature) and eventually filled in (death of the $H_1$ feature). From the resulting persistence diagram, we summarize the topology by the total \textit{lifetimes} of features in each dimension. Let $b_i$ and $d_i$ denote the birth and death radii of the $i$-th topological feature. We compute 
\[ 
\mathrm{Life}_0 = \sum_{i \in H_0} (d_i - b_i), 
\qquad 
\mathrm{Life}_1 = \sum_{i \in H_1} (d_i - b_i). 
\] 
Here, the sum for $H_0$ excludes the one infinite component that spans the whole class, and similarly for $H_1$ we include only loops that eventually get filled. Intuitively, a smaller total $\mathrm{Life}_0$ means that most points joined a single cluster at a small radius (i.e. the class manifold is tightly clustered), whereas a larger $\mathrm{Life}_0$ indicates more persistent disconnected clusters (a more fragmented manifold). $\mathrm{Life}_1$ captures the prevalence of persistent loops or cycles in the data. We compute these summaries for each class and average them to obtain an overall $\mathrm{Life}_0$ and $\mathrm{Life}_1$ for the model. In our analysis, PH serves as a topology-only baseline: it captures some notion of cluster tightness and loopiness, but unlike our weighted graph metrics, it ignores the graded strength of connections and often fails to distinguish finer geometric differences. By comparing the predictive power of PH features against spectral torsion or curvature, we assess whether incorporating geometric weighting yields stronger signals of robustness. Figure~\ref{fig:ph_ood} visualizes the relationship between the torsion proxy and persistent homology summaries across checkpoints, with color indicating OOD accuracy. While higher OOD accuracy aligns clearly with lower torsion (reduced spectral complexity), the corresponding PH signals exhibit greater dispersion, reinforcing that PH captures complementary but less predictive structure than the proposed geometric diagnostics.

\section{Extended Theoretical Perspective}
\label{app:theory}

We provide additional discussion connecting representation geometry to robustness and generalization.

While our study is primarily empirical, the observed relationship between representation geometry and out-of-distribution (OOD) robustness admits a natural interpretation through stability, smoothness, and margin-based views of generalization.

From a perturbation perspective, robustness under distribution shift requires that small input corruptions do not induce large displacements in representation space relative to class boundaries. Let $z = f_\theta(x)$ denote the embedding of an input $x$. A necessary condition for stable prediction is that perturbations $\delta$ satisfy
\begin{equation}
\| f_\theta(x+\delta) - f_\theta(x) \| \ll \mathrm{dist}(z, \mathcal{B}),
\end{equation}
where $\mathcal{B}$ denotes the decision boundary in representation space. This condition links robustness to the local smoothness and global organization of embeddings.

Mean Ollivier--Ricci curvature provides a proxy for local stability. Positive or weakly negative curvature indicates that neighboring distributions contract under optimal transport, implying that local perturbations tend to preserve nearest-neighbor relations and class-specific connectivity. As a result, small corruptions are less likely to push samples across unstable regions of the representation manifold, yielding smoother and more redundant neighborhoods that support stable classification.

The torsion proxy captures complementary global structure. The reduced log-determinant of the normalized Laplacian aggregates the full graph spectrum and reflects the redundancy and cyclic complexity of within-class connectivity. High torsion corresponds to highly entangled, cycle-rich graphs with multiple unstable pathways, amplifying the effect of perturbations. In contrast, lower torsion reflects simpler, more coherent class manifolds with fewer unstable directions.

This interpretation aligns with margin- and flatness-based views of generalization. Lower spectral complexity and higher local curvature jointly imply that embeddings concentrate in well-connected regions with larger effective margins between classes. From this perspective, curvature relates to local Lipschitz continuity of the embedding map, while torsion reflects global capacity and redundancy of the induced graph. Models with low torsion and high curvature therefore realize representations that are simultaneously smooth and globally regularized.

Our findings are also consistent with connections between robustness, manifold regularity, and feature contraction. Representation graphs with high curvature and low torsion resemble discretizations of manifolds with bounded curvature and low topological complexity, which support stable transport and diffusion dynamics. In contrast, highly tangled graphs exhibit unstable diffusion behavior and increased sensitivity to local disruptions.

Finally, this perspective explains the sensitivity of our metrics to structure-breaking controls. Label and feature shuffling, as well as degree-preserving rewiring, destroy neighborhood coherence and cyclic structure, disrupting both contraction and global regularity. Consequently, the geometric proxies lose predictive power, whereas isometry-preserving transformations retain strong correlations with robustness.

Taken together, torsion and Ricci curvature act as complementary proxies for two fundamental requirements of robust representation learning: global regularization of class manifolds and local stability of neighborhood relations. While not providing formal guarantees, this perspective offers a principled explanation for why representation geometry serves as an effective diagnostic of OOD robustness.

\section{Additional Experimental Results}
\label{app:additional}

\subsection{Multi-scale heat traces and spectral descriptors}
\label{app:heat}
Beyond torsion, we consider additional spectral invariants to capture the graph structure at multiple scales. First, the \textbf{algebraic connectivity} $\lambda_2(G_c)$, defined as the second-smallest eigenvalue of $L_{\mathrm{sym}}$, provides a measure of overall graph connectivity or the size of the toughest cut (higher $\lambda_2$ indicates the graph is well-connected and has no easily separated clusters). We also compute the \textbf{spectral entropy} of the Laplacian eigenvalue distribution as $-\sum_i \tilde{\lambda}_i \log \tilde{\lambda}_i$, where $\tilde{\lambda}_i = \lambda_i / \sum_j \lambda_j$ (normalizing the eigenvalues to sum to 1). This entropy measures how evenly spread out the spectrum is (low entropy indicates one or a few dominant eigenvalues, whereas high entropy indicates a flatter spectrum). 

Most prominently, we examine the \textbf{heat kernel trace} at multiple time scales $t$. The heat trace $H_c(t)$ for class $c$ is defined as the trace of the heat kernel $\exp(-t L_{\mathrm{sym}})$, which can be expressed in terms of the eigenvalues as:
\[ 
H_c(t) = \mathrm{Tr}\!\big(\exp(-tL_{\mathrm{sym}})\big) \;=\; \sum_{i} \exp(-t\,\lambda_i)\,. 
\] 
This quantity effectively counts a weighted number of closed walks of length related to $t$ on the graph, thus capturing connectivity at different scales: small $t$ probes very local connectivity (since $\exp(-t\lambda_i) \approx 1$ for all small eigenvalues, counting even small clusters), while large $t$ emphasizes global connectivity (heavily damping all but the smallest eigenvalues, thus sensitive to large-scale structure like whether the graph is close to bipartite or has many components). We evaluate $H_c(t)$ at several values of $t$ (e.g., $t \in \{0.1,\,0.3,\,1.0,\,3.0\}$) for each class graph, and either analyze specific scales or perform a principal component analysis on the vector $(H_c(0.1), \dots, H_c(3.0))$ to obtain a single composite descriptor. As with torsion, we average the heat trace (or its derived features) over classes to produce one multiscale connectivity profile per model. These spectral descriptors provide a fingerprint of each class manifold’s structure and allow us to examine which structural scale (local vs. global) most strongly correlates with robustness.

%%%%%%%%%%%%%%%%%%%%%%%%%%%%%%%%%%%%%%%%%%%%%%%%%%%%%%%%%%%%%%%%%%%%%%%%%%%%%%%

\subsection{Ablation and Per-Class Analysis}
\label{app:ablations}
Here we want to ask whether the geometry--robustness signal is uniform across all classes or driven by a subset of classes. To investigate this, we compute class-wise versions of our metrics and correlate them with class-specific OOD accuracies. Figure~\ref{fig:perclass} in the appendix presents the per-class correlation results. The analysis reveals substantial heterogeneity: for some classes, the geometric metrics are strongly predictive of that class’s OOD performance, whereas for others the correlation is weaker. For example, classes that benefit most from learned geometric structure (e.g. those with many visually diverse samples) show a pronounced increase in local Ricci curvature and decrease in torsion as their OOD accuracy improves. In contrast, for classes that remain easy even under shift, geometric changes are less tied to performance. This per-class variation highlights that our global results are not dominated by any single class, yet certain challenging classes contribute disproportionately to the observed geometry–accuracy coupling. It also underscores the value of class-conditional analysis: aggregating over classes could obscure these fine-grained patterns where robustness is most tightly linked to representation geometry.

We further assess the robustness of our findings under different analysis choices.
Using weighted $k$-NN graphs instead of unweighted edges slightly improves correlation
stability, suggesting that incorporating similarity magnitudes is beneficial. Varying
the number of neighbors ($k = 5, 10, 15$), extracting embeddings from different layers
(e.g., pre-classifier versus penultimate features), or modifying preprocessing
(PCA-whitening versus $\ell_2$-normalization) does not qualitatively affect the results.
Across all settings, torsion and curvature remain predictive of OOD robustness. Detail ablation results are reported in~\ref{tab:ablations2}.

\begin{figure*}[t]
\centering
\includegraphics[width=0.9\linewidth]{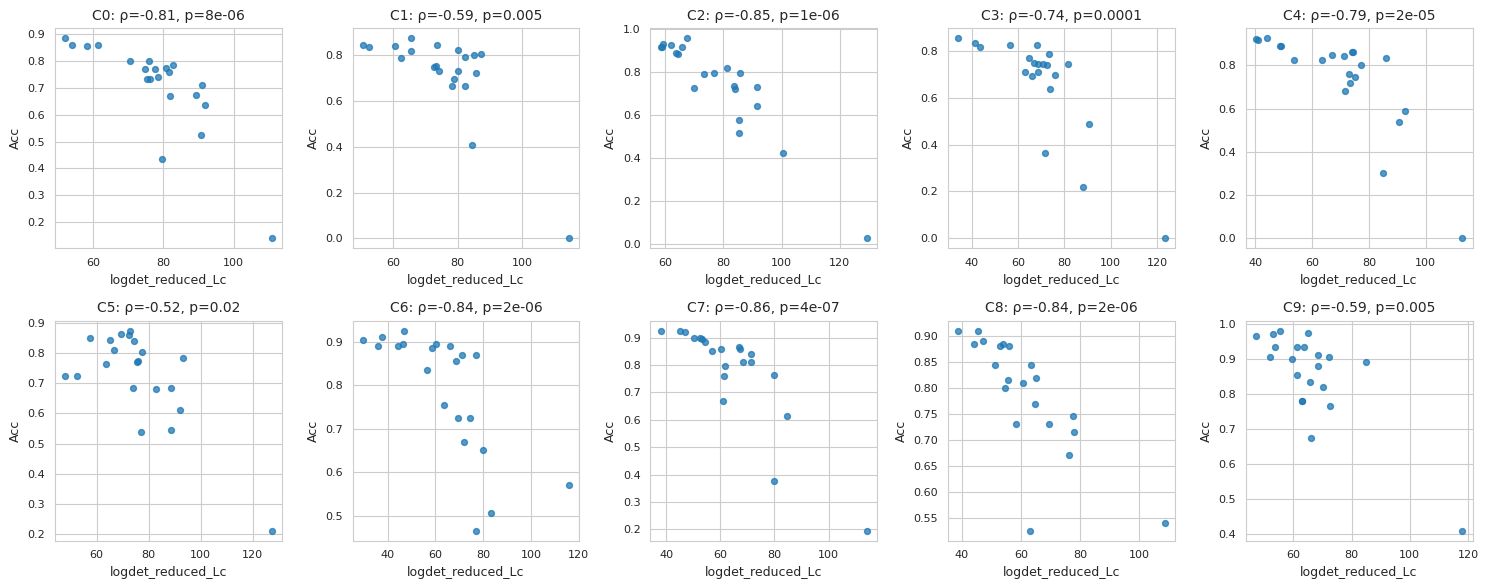}
\caption{Torsion proxy vs OOD accuracy.}
\label{fig:perclass}
\end{figure*}

\begin{table*}[t]
\centering
\caption{Layer- and $k$-sweep summary (mean $\pm$ std across runs/bootstraps).}
\label{tab:ablations2}
\begin{tabular}{llrrrrrr}
\toprule
\textbf{Layer} & \textbf{$k$} &
$\log\det^* L_C$ & Entropy & $\lambda_2(L_C)$ & $h(0.1)$ & $H_0$ Life & $\bar{\kappa}$ (mean Ricci) \\
\midrule
avgpool        & 5  & -235.444199 & 4.977678 & 3.424739e-16 & 251.929546 & 468.239393 &  0.042421 \\
avgpool        & 10 &  498.700192 & 5.298542 & 1.293900e-16 & 247.559453 & 468.239393 & -0.111212 \\
avgpool        & 15 & 1002.215896 & 5.372179 & 3.710285e-16 & 244.966772 & 468.239393 & -0.133393 \\
\midrule
preclassifier  & 5  & -313.553671 & 4.996368 & 8.604969e-17 & 251.661174 & 718.994737 &  0.058089 \\
preclassifier  & 10 &  330.293156 & 5.289406 & 2.320415e-16 & 247.727011 & 718.994737 & -0.106870 \\
preclassifier  & 15 &  845.126823 & 5.377014 & 4.488768e-16 & 245.029390 & 718.994737 & -0.123253 \\
\bottomrule
\end{tabular}
\end{table*}

\section{Mechanism and falsifiable hypotheses}
\label{sec:mechanism}
OOD robustness is fundamentally about \emph{stability of class identity under perturbations}. In representation space, a corruption acts like a perturbation that should not move a sample outside its within-class neighborhood. This motivates two complementary geometric desiderata on class-conditional representation graphs.

\textbf{Local contraction and smoothness: }
If within-class neighborhoods are coherent, small perturbations preserve local nearest-neighbor relations. Ollivier--Ricci curvature is positive when neighborhoods \emph{contract} under optimal transport, reflecting locally regular geometry. \emph{Hypothesis H1:} robust checkpoints increase within-class local contraction, hence exhibit higher mean curvature $\bar{\kappa}(\theta)$.

\textbf{Reduced global tangling: }
Global spectral complexity measures how many independent modes of variation the graph supports. The reduced $\log\det^*$ of the normalized Laplacian increases with connectivity redundancy and cycle-rich structure (many alternative paths), i.e., a more tangled within-class graph. \emph{Hypothesis H2:} robust checkpoints reduce cycle-rich tangling within classes, hence exhibit lower torsion proxy $\mathcal{T}(\theta)$.

\textbf{Why this torsion proxy and why class-conditional graphs: }
We use $\log\det^*(L_{\mathrm{sym}})$ as a principled global proxy that aggregates the entire Laplacian spectrum (unlike single-eigenvalue summaries such as $\lambda_2$) and has a classical combinatorial interpretation via Kirchhoff’s Matrix--Tree theorem. Class-conditional graphs are essential because the hypotheses concern \emph{intra-class stability}: global graphs can conflate robustness with between-class separation and are dominated by cross-class distances rather than within-class organization.

\textbf{Falsifiability via structure-breaking ablations: }
The mechanism predicts the signal should \emph{attenuate} when task-aligned geometry is destroyed. \emph{H3:} label shuffling or feature shuffling should drive correlations toward $0$ because class semantics and neighborhood structure are broken. \emph{H4:} degree-preserving edge rewiring should weaken torsion/curvature associations because cycles and local neighborhoods are disrupted while trivial graph statistics (size/degree) are preserved.

\end{document}